\documentclass{article}
\usepackage{spconf,amsmath,graphicx}

\usepackage{amssymb,amsfonts,bm}  
\usepackage{color}  
\usepackage[hyperindex,breaklinks]{hyperref} 
\usepackage{makecell} 

\title{PARALLEL-DATA-FREE VOICE CONVERSION\\
  USING CYCLE-CONSISTENT ADVERSARIAL NETWORKS}
\name{Takuhiro Kaneko, Hirokazu Kameoka}
\address{NTT Communication Science Laboratories, NTT Corporation, Japan}
\begin{document}
%
\maketitle
\begin{abstract}
  We propose a parallel-data-free voice-conversion (VC) method that can learn a mapping from source to target speech without relying on parallel data. The proposed method is general purpose, high quality, and parallel-data free and works without any extra data, modules, or alignment procedure. It also avoids over-smoothing, which occurs in many conventional statistical model-based VC methods. Our method, called CycleGAN-VC, uses a cycle-consistent adversarial network (CycleGAN) with gated convolutional neural networks (CNNs) and an identity-mapping loss. A CycleGAN learns forward and inverse mappings simultaneously using adversarial and cycle-consistency losses. This makes it possible to find an optimal pseudo pair from unpaired data. Furthermore, the adversarial loss contributes to reducing over-smoothing of the converted feature sequence. We configure a CycleGAN with gated CNNs and train it with an identity-mapping loss. This allows the mapping function to capture sequential and hierarchical structures while preserving linguistic information. We evaluated our method on a parallel-data-free VC task. An objective evaluation showed that the converted feature sequence was near natural in terms of global variance and modulation spectra. A subjective evaluation showed that the quality of the converted speech was comparable to that obtained with a Gaussian mixture model-based method under advantageous conditions with parallel and twice the amount of data.
\end{abstract}
\begin{keywords}
  Voice conversion, parallel-data free, generative adversarial networks,
  CycleGAN, gated CNN
\end{keywords}
\section{Introduction}
\label{sec:introduction}

Voice conversion (VC) is a technique to modify
non/para-linguistic information of speech
while preserving linguistic information.
This technique can be applied to various tasks such as
speaker-identity modification for text-to-speech (TTS)
systems~\cite{AKainICASSP1998},
speaking assistance~\cite{AKainSC2007,KNakamuraSC2012},
speech enhancement~\cite{ZInanogluSC2009,OTurkTASLP2010,TTodaTASLP2012},
and pronunciation conversion~\cite{TKanekoIS2017b}.

Voice conversion can be formulated as a regression problem
of estimating a mapping function from source to target speech.
One successful approach involves statistical methods using
a Gaussian mixture model (GMM)~\cite{YStylianouTASP1998,TTodaTASLP2007,EHelanderTASLP2010}.
Neural network (NN)-based methods,
such as
a restricted Boltzmann machine (RBM)~\cite{LChenTASLP2014,TNakashikaIEICE2014},
feed forward NN~\cite{SDesaiTASLP2010,SMohammadiSLT2014},
recurrent NN (RNN)~\cite{TNakashikaIS2014,LSunICASSP2015},
and convolutional NN (CNN)~\cite{TKanekoIS2017b},
and exemplar-based methods,
such as non-negative matrix factorization
(NMF)~\cite{RTakashimaIEICE2013,ZWuTASLP2014},
have also recently been proposed.

Many VC methods including those mentioned above
typically use temporally aligned parallel data of
source and target speech as training data.
If perfectly aligned parallel data are available,
obtaining the mapping function becomes relatively simple;
however,
collecting such data can be a painstaking process in real application scenarios.
Even though we could collect such data,
we need to perform automatic time alignment,
which may occasionally fail.
This can be problematic since misalignment involved in parallel data
can cause speech-quality degradation; thus,
careful pre-screening and manual correction may be required
~\cite{EHelanderIS2008}.

These facts motivated us to consider a VC problem
that is free from parallel data.
In this paper, we propose a parallel-data-free VC method,
which is particularly noteworthy in that it
(1) does not require any extra data, such as
transcripts and reference speech,
and extra modules, such as an automatic speech-recognition (ASR) module,
(2) is not prone to over-smoothing, which is known to be one of the main factors
leading to speech-quality degradation, and
(3) captures a spectrotemporal structure without any alignment procedure.

To satisfy these requirements,
our  method, called
\sloppy
CycleGAN-VC,
\sloppy
uses a cycle-consistent adversarial network (CycleGAN)~\cite{JYZhuICCV2017}
(i.e., DiscoGAN~\cite{TKimICML2017} or DualGAN~\cite{ZYiICCV2017})
with gated CNNs~\cite{YDauphinICML2017} and
an identity-mapping loss~\cite{YTaigmanICLR2017}.
The CycleGAN was originally proposed
for unpaired image-to-image translation.
With this model, forward and inverse mappings are simultaneously learned
using an adversarial loss~\cite{IGoodfellowNIPS2014}
and cycle-consistency loss~\cite{TZhouCVPR2016}.
This makes it possible to find
an optimal pseudo pair from unpaired data.
Furthermore,
the adversarial loss does not require explicit density estimation
and results in reducing the over-smoothing
effect~\cite{TKanekoIS2017b,TKanekoICASSP2017,TKanekoIS2017a,YSaitoICASSP2017}.
To use a CycleGAN for parallel-data-free VC,
we configure a network using gated CNNs
and train it with an identity-mapping loss.
This allows the mapping function
to capture sequential and hierarchical structures
while preserving linguistic information.

We evaluated our method on a parallel-data-free VC task
using the Voice Conversion Challenge 2016 (VCC 2016) dataset~\cite{VCC2016}.
An objective evaluation showed that
the converted feature sequence
was reasonably good
in terms of global variance (GV)~\cite{TTodaTASLP2007} and
modulation spectra (MS)~\cite{STakamichiICASSP2014}.
A subjective evaluation showed that
the speech quality was comparable
to that obtained with
a GMM-based method~\cite{TTodaTASLP2007}
trained using parallel and twice the amount of data.
This is noteworthy
since our method had a disadvantage in the training condition.

This paper is organized as follows.
In Section~\ref{sec:related}, we describe related work.
In Section~\ref{sec:method}, we review the CycleGAN and
explain our proposed method (CycleGAN-VC).
In Section~\ref{sec:experiments}, we report on the experimental results.
In Section~\ref{sec:conclusions},
we provide a discussion and conclude the paper.

\section{Related work}
\label{sec:related}
Recently, several approaches for parallel-data-free VC have been proposed.
One approach involves using
an ASR module
to find a pair of corresponding frames~\cite{MZhangICASSP2008,FLXieIS2016}.
This may work well if ASR performs robustly and accurately enough,
but it requires a large amount of transcripts to train the ASR module.
It would also be inherently difficult to capture nonverbal information.
This may become a limitation to be applied
in general situations.
Other approaches involve methods using an adaptation
technique~\cite{AMouchtarisTASLP2006,CHLeeIS2006}
or incorporating a pre-constructed speaker space~\cite{TTodaIS2006,DSaitoIS2011}.
These methods do not require parallel data
between source and target speakers
but require parallel data among reference speakers.
A few attempts~\cite{TNakashikaTASLP2016,CHsuAPSIPA2016,TKinnunenICASSP2017,CHsuIS2017} have recently been made to develop methods
that are completely free from parallel data and extra modules.
With these methods, it is assumed that source and target speech
lie in the same low-dimensional embeddings.
This would not only limit applications but also
cause difficulty in modeling complex structures,
e.g., detailed spectrotemporal structure.
In contrast, we learn a mapping function directly without embedding.
We expect that this would make it possible
to apply our method to various applications
where complex structure modeling needs to be considered.


\section{Parallel-data-free VC using CycleGAN}
\label{sec:method}

\subsection{CycleGAN}

Our goal is to learn a mapping from source $x \in X$ to target $y \in Y$
without relying on parallel data.
We solve this problem based on a CycleGAN~\cite{JYZhuICCV2017}.
In this subsection, we briefly review the concept of CycleGAN and
in the next subsection,
we explain our proposed method for parallel-data-free VC.

\begin{figure}[t]
  \centerline{\includegraphics[width=\columnwidth]{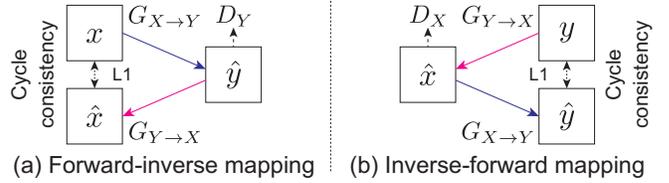}}
  \vspace{-4mm}
  \caption{Training procedure of CycleGAN}
  \vspace{-5mm}
  \label{fig:cyclegan}
\end{figure}

With CycleGAN, a mapping $G_{X \rightarrow Y}$ is learned using two losses,
namely an adversarial loss~\cite{IGoodfellowNIPS2014} and
cycle-consistency loss~\cite{TZhouCVPR2016}.
We illustrate the training procedure in Fig.~\ref{fig:cyclegan}.

{\bf Adversarial loss:}
An adversarial loss measures
how distinguishable converted data $G_{X \rightarrow Y}(x)$
are from target data~$y$.
Hence, the closer the distribution of converted data
$P_{G_{X \rightarrow Y}}(x)$
becomes to that of target data  $P_{\rm Data}(y)$,
the smaller this loss becomes.
This objective is written as
\begin{align}
  \label{eqn:gan}
        {\cal L}_{adv}(G_{X \rightarrow Y}, D_Y)
        = {\mathbb E}_{y \sim P_{\rm Data}(y)}
        [\log D_Y(y)]\:\:\:\:\:\:\:\:\:\:\:\:
        \nonumber \\
        + {\mathbb E}_{x \sim P_{\rm Data}(x)}
        [\log(1 - D_Y(G_{X \rightarrow Y}(x)))].
\end{align}
The generator $G_{X\rightarrow Y}$ attempts to generate data
indistinguishable from target data $y$ by the discriminator $D_Y$
by minimizing this loss,
whereas $D_Y$ attempts not to be deceived
by $G_{X\rightarrow Y}$ by maximizing this loss.

{\bf Cycle-consistency loss:}
Optimizing only the adversarial loss would not necessarily guarantee
that the contextual information of $x$ and
$G_{X\rightarrow Y}(x)$ will be consistent.
This is because the adversarial loss only tells us
whether $G_{X\rightarrow Y}(x)$ follows the target-data distribution and
does not help preserve the contextual information of $x$.
The idea of CycleGAN \cite{JYZhuICCV2017} is
to introduce two additional terms.
One is an adversarial loss ${\cal L}_{adv}(G_{Y \rightarrow X}, D_X)$
for an inverse mapping $G_{Y\rightarrow X}$ and
the other is a cycle-consistency loss,
given as
\begin{align}
  \label{eqn:cycle}
        & {\cal L}_{cyc}(G_{X \rightarrow Y}, G_{Y \rightarrow X})
        \nonumber \\
        = \: & {\mathbb E}_{x \sim P_{\rm Data}(x)}
        [|| G_{Y \rightarrow X}(G_{X \rightarrow Y}(x)) - x ||_1]
        \nonumber \\
        + \: & {\mathbb E}_{y \sim P_{\rm Data}(y)}
        [|| G_{X \rightarrow Y}(G_{Y \rightarrow X}(y)) - y ||_1].
\end{align}
These additional terms encourage
$G_{X \rightarrow Y}$ and $G_{Y \rightarrow X}$
to find $(x, y)$ pairs with the same contextual information.

The full objective is written with trade-off parameter $\lambda_{cyc}$:
\begin{align}
  \label{eqn:full}
        {\cal L}_{full}
        = \: & {\cal L}_{adv}(G_{X \rightarrow Y}, D_Y)
        + {\cal L}_{adv}(G_{Y \rightarrow X}, D_X)
        \nonumber \\
        + \: & \lambda_{cyc}
        {\cal L}_{cyc}(G_{X \rightarrow Y}, G_{Y \rightarrow X}).
\end{align}

\begin{figure*}[t]
  \centerline{\includegraphics[width=\textwidth]{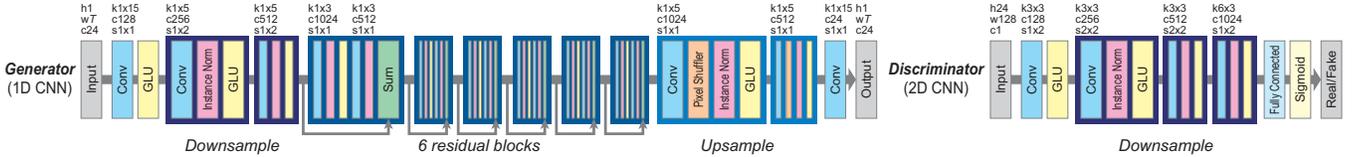}}
  \vspace{-4.5mm}
  \caption{Network architectures of generator and discriminator.
    In input or output layer,
    {\tt h}, {\tt w}, and {\tt c} represent
    height, width, and number of channels, respectively.
    In each convolutional layer,
    {\tt k}, {\tt c}, and {\tt s} denote
    kernel size, number of channels, and stride size, respectively.
    Since generator is fully convolutional~\cite{JLongCVPR2015},
    it can take input of arbitrary length $T$.}
  \label{fig:network}
  \vspace{-6mm}
\end{figure*}

\vspace{-3mm}

\subsection{CycleGAN for parallel-data-free VC: CycleGAN-VC}
\label{subsec:vc-cyclegan}

To use a CycleGAN for parallel-data-free VC,
we mainly made two modifications to the CycleGAN architecture:
gated CNN~\cite{YDauphinICML2017}
and identity-mapping loss~\cite{YTaigmanICLR2017}.

{\bf Gated CNN:}
One of the characteristics of speech is that
it has sequential and hierarchical structures,
e.g., voiced/unvoiced segments and phonemes/morphemes.
An effective way to represent such structures would be to use an RNN,
but it is computationally demanding
due to the difficulty of parallel implementations.
Instead, we configure a CycleGAN using
gated CNNs~\cite{YDauphinICML2017},
which not only allows parallelization over sequential data
but also achieves state-of-the-art in
language modeling~\cite{YDauphinICML2017}
and speech modeling~\cite{TKanekoIS2017b}.
In a gated CNN, gated linear units (GLUs) are used as an activation function.
A GLU is a data-driven activation function, and
the $(l+1)$-th layer output $\bm{H}_{l+1}$ is
calculated using the $l$-th layer output $\bm{H}_l$ and
model parameters $\bm{W}_l$, $\bm{V}_l$, $\bm{b}_l$, and~$\bm{c}_l$,
\begin{equation}
  \label{eqn:glu}
  \bm{H}_{l+1} = (\bm{H}_l * \bm{W}_l + \bm{b}_l)
  \otimes \sigma (\bm{H}_l * \bm{V}_l + \bm{c}_l),
\end{equation}
where $\otimes$ is the element-wise product and
$\sigma$ is the sigmoid function.
This gated mechanism allows the information to be selectively propagated
depending on the previous layer states.

{\bf Identity-mapping loss:}
A cycle-consistency loss provides constraints on a structure;
however, it would not suffice to guarantee that
the mappings always preserve linguistic information.
To encourage linguistic-information preservation
without relying on extra modules, 
we incorporate an identity-mapping loss~\cite{YTaigmanICLR2017},
\begin{eqnarray}
  \label{eqn:idt}
        {\cal L}_{id} (G_{X \rightarrow Y}, G_{Y \rightarrow X})
        = {\mathbb E}_{y \sim P_{\rm Data}(y)}
        [|| G_{X \rightarrow Y}(y) - y||_1]
        \nonumber \\
        + \: {\mathbb E}_{x \sim P_{\rm Data}(x)}
        [|| G_{Y \rightarrow X}(x) - x||_1],
\end{eqnarray}
which encourages the generator
to find the mapping that preserves composition
between the input and output.
In practice, weighted loss $\lambda_{id} {\cal L}_{id}$ with
trade-off parameter $\lambda_{id}$ is added to
Eq.~\ref{eqn:full}.
Note that the original study on CycleGANs~\cite{JYZhuICCV2017}
showed the effectiveness of this loss for color preservation.


\section{Experiments}
\label{sec:experiments}

\vspace{-1mm}
\subsection{Experimental conditions}
\label{subsec:exp_cond}

We conducted experiments to evaluate our method
on a parallel-data-free VC task.
We used the VCC 2016 dataset~\cite{VCC2016},
which was recorded by professional US English speakers,
including five females and five males.
Following a previous study~\cite{CHsuAPSIPA2016},
we used a subset of speakers for evaluation.
A pair of female (SF1) and male (SM1) speakers were selected as sources and
another pair (TF2 and TM3) were selected as targets.
The audio files for each speaker were manually segmented into
216 short parallel sentences (about 13 minutes).
Among them,
162 and 54 sentences were provided as training and evaluation sets,
respectively.
To evaluate our method under a parallel-data-free condition,
we divided the training set into two subsets without overlap.
For the first half, 81 sentences were used for the source and
the other 81 sentences were used for the target.
The speech data were downsampled to 16~kHz, and
24 Mel-cepstral coefficients (MCEPs),
logarithmic fundamental frequency ($\log F_0$), and
aperiodicities (APs) were then extracted every 5~ms
using the WORLD analysis system~\cite{MMoriseIEICE2016}.
Among these features,
we learned a mapping in the MCEP domain using our method.
The $F_0$ was converted using logarithm Gaussian normalized transformation
\cite{KLiuFSKD2007}.
Aperiodicities were directly used without modification
because a previous study~\cite{YOhtaniIS2006}
showed that converting APs does not significantly affect speech quality.

{\bf Implementation details:}
We designed a network based on the recent success in
image modeling~\cite{JYZhuICCV2017,WShiCVPR2016,JJohnsonECCV2016}
and speech modeling~\cite{TKanekoIS2017b,TKanekoICASSP2017}.
The network architecture is illustrated in Fig.~\ref{fig:network}.
We designed the generator using a one-dimensional (1D) CNN~\cite{TKanekoIS2017b}
to capture the relationship among the overall features
while preserving the temporal structure.
Inspired by a previous study~\cite{JJohnsonECCV2016}
for neural style transfer and super-resolution,
we used the network that included
downsampling, residual~\cite{KHeCVPR2016}, and upsampling layers,
as well as incorporating instance normalization~\cite{DUlyanovArXiv2016}.
We used pixel shuffler
for upsampling,
which is effective for
high-resolution image generation~\cite{WShiCVPR2016}.
We designed the discriminator using a 2D CNN~\cite{TKanekoIS2017b}
to focus on a 2D spectral texture~\cite{TKanekoICASSP2017}.

\begin{figure}[t]
  \centerline{\includegraphics[width=0.97\columnwidth]{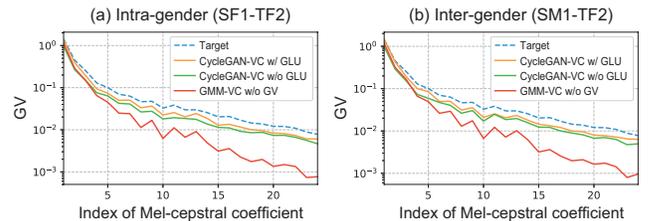}}
  \vspace{-4mm}
  \caption{Comparison of GV per MCEP$^1$}
  \vspace{-3mm}
  \label{fig:gv}
\end{figure}

\footnotetext[1]{
  We omit {\bf GMM-VC w/ GV} in Fig.~\ref{fig:gv}
  because it directly estimates GV.
}

\begin{figure}[t]
  \centerline{\includegraphics[width=0.97\columnwidth]{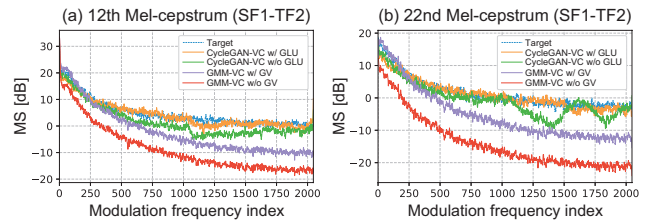}}
  \vspace{-4mm}
  \caption{Comparison of MS per modulation frequency}
  \vspace{-6mm}
  \label{fig:ms}
\end{figure}

\begin{table}[t]  
  \centering
  \vspace{-3mm}
  \caption{Comparison of RMSE between target and converted logarithmic MS
    averaged over all MCEPs and modulation frequencies [dB].
    Small value indicates closeness to target.}
  \label{tab:ms}
  \scriptsize{
    \begin{tabular}{lcccc}
      {\bf Method} & \!{\bf SF1--TF2}\! & \!{\bf SF1--TM3}\!
      & \!{\bf SM1--TF2}\! & \!{\bf SM1--TM3}\! \\
      \hline
      CycleGAN-VC w/ GLU  & {\bf 1.98} & {\bf 2.69} & {\bf 1.93} & {\bf 2.14} \\
      CycleGAN-VC w/o GLU &      3.34  &      2.99  &      3.17 &       2.94  \\
      GMM-VC w/ GV        &      7.59  &      9.41  &      8.69 &       9.67  \\
      GMM-VC w/o GV       &     13.56  &     14.90  &     14.17 &      14.53  \\
    \end{tabular}
  }
  \vspace{-4mm}
\end{table}

{\bf Training details:}
As a pre-process,
we normalized the source and target MCEPs
per dimension.
To stabilize training,
we used a least squares GAN~\cite{XMaoArXiv2016},
which replaces the negative log likelihood objective in ${\cal L}_{adv}$ by
a least squares loss.
We set $\lambda_{cyc} = 10$.
We used ${\cal L}_{id}$ only for the first $10^4$ iterations
with $\lambda_{id} = 5$ to guide the learning process.
To increase the randomness of each batch,
we did not use a sequence directly and
cropped a fixed-length segment (128 frames) randomly
from a randomly selected audio file.
We trained the network using the Adam optimizer~\cite{DPKingmaICLR2015}
with a batch size of $1$.
We set the initial learning rates to $0.0002$ for the generator
and $0.0001$ for the discriminator.
We kept the same learning rate for the first $2 \times 10^5$ iterations
and linearly decay over the next $2 \times 10^5$ iterations.
We set the momentum term $\beta_1$ to $0.5$.

\vspace{-0.9mm}
\subsection{Objective evaluation}
\label{subsec:exp_obj_eval}

In these experiments, we focused on the conversion of MCEPs;
therefore, we evaluated the quality of converted MCEPs.
We compared our method ({\bf CycleGAN-VC})
with a GMM-based method ({\bf GMM-VC})~\cite{TTodaTASLP2007}
because it is still comparable to a DNN-based method
in a relatively small dataset~\cite{MWesterSSW2016}.
Since this method requires parallel data,
all the training data (162 sentences) for both source and target
were used.
This means that the amount of training data was twice as ours.
As an ablation study, we examined our method without any GLUs.
Instead of GLUs, we used typical GAN activation functions,
i.e., rectified linear units (ReLUs)~\cite{VNairICML2010} for the generator and
leaky ReLUs~\cite{AMaasICML2013,BXuICMLW2015} for the discriminator.
In the pre-experiment,
we also examined our method without an identity-mapping loss.
This revealed that the lack of this loss tends to cause significant degradation,
e.g., collapse of the linguistic structure;
thus, we did not examine this further.

\begin{figure}[t]
  \centerline{\includegraphics[width=\columnwidth]{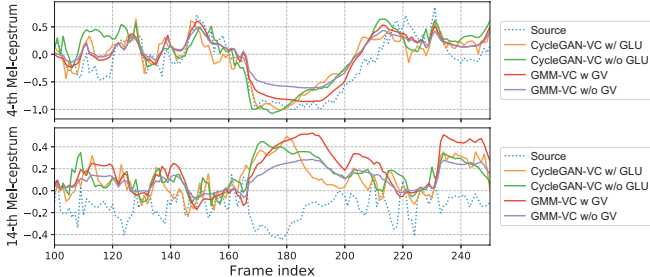}}
  \vspace{-4mm}
  \caption{Comparison of MCEP trajectories (SF1-TM3)}
  \vspace{-6mm}
  \label{fig:traj}
\end{figure}

Mel-cepstral distortion is a well-used measure to evaluate the quality of
synthesized MCEPs,
but recent studies~\cite{TTodaTASLP2007,TKanekoICASSP2017,CHsuIS2017}
indicate the limitation of this measure:
it tends to prefer over-smoothing
because it internally assumes Gaussian distribution.
Therefore, as alternatives, we used two structural indicators
highly correlated with subjective evaluation:
GV~\cite{TTodaTASLP2007}
and MS~\cite{STakamichiICASSP2014}.
We show the comparison of GV in Fig.~\ref{fig:gv}.
We list the comparison of root mean squared error (RMSE)
between target and converted logarithmic MS in Table~\ref{tab:ms}.
We also show the comparison of MS
per modulation frequency in Fig.~\ref{fig:ms}.
These results indicate that the MCEP sequences obtained with our method
({\bf CycleGAN-VC w/ GLU})
are closest to the target in terms of GV and MS.
We expect this is because
(1) the adversarial loss does not require
explicit density estimation;
thus, avoids over-smoothing,
and
(2) the GLU is a data-driven activation function; therefore,
it can represent sequential and hierarchical structures better
than the ReLU and leaky ReLU.
We show sample MCEP trajectories in Fig.~\ref{fig:traj}.
The trajectories of {\bf CycleGAN-VC w/ GLU} have
a similar global structure to those of {\bf GMM-VC w/ GV}
while preserving similar complexity to the source.



\begin{figure}[t]
  \centerline{\includegraphics[width=0.8\columnwidth]{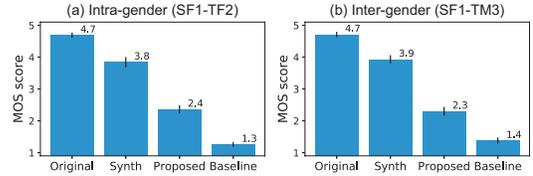}}
  \vspace{-4mm}
  \caption{MOS for naturalness with 95\% confidence intervals}
  \vspace{-3.5mm}
  \label{fig:mos}
\end{figure}

\begin{figure}[t]
  \centerline{\includegraphics[width=\columnwidth]{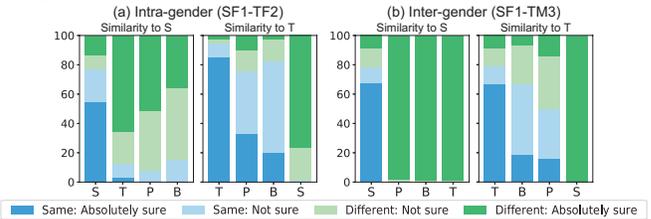}}
  \vspace{-4mm}
  \caption{Similarity to source speaker and to target speaker
    (S: Source, T: Target, P: Proposed, and B: Baseline)}
  \vspace{-6mm}
  \label{fig:xa}
\end{figure}

\vspace{-0.6mm}
\subsection{Subjective evaluation}
\label{subsec:exp_subj_eval}

We conducted listening tests to evaluate the performance of converted
speech\footnote[2]{We provide the converted speech samples at
  \href{http://www.kecl.ntt.co.jp/people/kaneko.takuhiro/projects/cyclegan-vc}{\tt http://www.kecl.}\\
  \href{http://www.kecl.ntt.co.jp/people/kaneko.takuhiro/projects/cyclegan-vc}{\tt ntt.co.jp/people/kaneko.takuhiro/projects/}\\
  \href{http://www.kecl.ntt.co.jp/people/kaneko.takuhiro/projects/cyclegan-vc}{\tt cyclegan-vc}}.
By referring to the VCC 2016~\cite{MWesterIS2016},
we evaluated the naturalness and speaker similarity of the converted samples.
We compared our method with the baseline of the VCC
2016\footnote[3]{We used data at
  \url{http://dx.doi.org/10.7488/ds/1575}},
which is a GMM-based method~\cite{TTodaTASLP2007}
using parallel and twice the amount of data.
To measure naturalness,
we conducted a mean opinion score (MOS) test.
As a reference, we used original
and synthesized-and-analyzed
(upper bound of our method) speeches of target speakers.
Twenty sentences longer than 2~s and shorter than 5~s were randomly selected
from the evaluation sets.
To measure speaker similarity,
we used the same/different paradigm~\cite{MWesterIS2016}.
Ten sample pairs were randomly selected from the evaluation sets.
There were nine participants who were well-educated English speakers.
By referring to the study by~\cite{CHsuIS2017},
we evaluated on two subsets: intra-gender VC (SF1--TF2)
and inter-gender VC (SF1--TM3).

We show the MOS for naturalness in Fig.~\ref{fig:mos}.
The results indicate that the proposed method significantly outperformed
the baseline.
We show the similarity to a source speaker and to a target speaker
in Fig.~\ref{fig:xa}.
The results indicate that our method was
slightly inferior to the baseline in SF1--TM3 VC but
superior in SF1--TF2 VC.
Overall, our method is comparable to the baseline.
This is noteworthy since our method is trained under disadvantageous conditions
with half the amount of and non-parallel data.

\vspace{-0.6mm}
\section{Discussion and conclusions}
\label{sec:conclusions}

We proposed a parallel-data-free VC method called CycleGAN-VC,
which uses a CycleGAN with gated CNNs and an identity-mapping loss.
This method can learn a sequence-based mapping function
without any extra data, modules, and time alignment procedure.
An objective evaluation showed that
the MCEP sequences obtained with our method are close to the target
in terms of GV and MS.
A subjective evaluation showed that the quality of converted speech was
comparable to that obtained with the GMM-based method
under advantageous conditions with parallel and twice the amount of data.
However, there is still a margin between original and converted speeches.
To fill the margin,
we plan to apply our method to other features, such as
STFT spectrograms~\cite{TKanekoIS2017a}, and
other speech-synthesis frameworks, such as
vocoder-free VC~\cite{KKobayashiSLT2016}.
Furthermore, our proposed method is a general framework, and
possible future work includes applying the method to 
other VC applications~\cite{
  AKainICASSP1998,
  AKainSC2007,KNakamuraSC2012,
  ZInanogluSC2009,OTurkTASLP2010,TTodaTASLP2012,
  TKanekoIS2017b}.


{\bf Acknowledgements:}
This work was supported by JSPS KAKENHI Grant Number 17H01763.

\newpage
\bibliographystyle{IEEEbib}
\scriptsize{\bibliography{refs}}

\end{document}